\documentclass{article}

\usepackage{color}
\usepackage{graphicx}
\usepackage{ulem}
\usepackage{hyperref}
\usepackage{subfigure}
\usepackage{amsthm,amssymb,amsbsy,amsmath,amsfonts,amssymb,amscd}

\newcommand{\E}{\mathrm{E}}
\newcommand{\V}{\mathrm{var}}
\newcommand{\C}{\mathrm{cov}}
\newcommand{\p}{\mathrm{P}}

\newtheorem{prop}{Proposition}

\usepackage[usenames,dvipsnames]{pstricks}
\usepackage{multirow}

\title{
Additive Kernels for Gaussian Process Modeling
}

\date{\null}

\begin{document}

\vspace{-1cm}

\maketitle

\vspace{-2cm}

\begin{center}
\Large
N. Durrande\footnotemark[1]\footnotemark[3], D. Ginsbourger\footnotemark[2], O. Roustant \footnotemark[1]\\
January 12, 2010
\end{center}

\footnotetext[1]{CROCUS - Ecole Nationale Sup\'erieure des Mines de St-Etienne 
        \\ \phantom{ha ha} 29 rue Ponchardier - 42023 St Etienne, France}
\footnotetext[2]{Institute of Mathematical Statistics and Actuarial Science, University of Berne,
        \\ \phantom{ha ha}  Alpeneggstrasse 22 - 3012 Bern, Switzerland,}
\footnotetext[3]{Corresponding author: \textit{durrande@emse.fr}}

\begin{abstract}
Gaussian Process (GP) models are often used as mathematical approximations of computationally expensive experiments.  
Provided that its kernel is suitably chosen and that enough data is available to obtain a reasonable fit of the simulator, a GP model can beneficially be used for tasks such as prediction, optimization, or Monte-Carlo-based quantification of uncertainty. However, the former conditions become unrealistic when using classical GPs as the dimension of input increases. One popular alternative is then to turn to Generalized Additive Models (GAMs), relying on the assumption that the simulator's response can approximately be decomposed as a sum of univariate functions. If such an approach has been successfully applied in approximation, it is nevertheless not completely compatible with the GP framework and its versatile applications.
The ambition of the present work is to give an insight into the use of GPs for additive models by integrating additivity within the kernel, and proposing a parsimonious numerical method for data-driven parameter estimation.
The first part of this article deals with the kernels naturally associated to additive processes and the properties of the GP models based on such kernels. 
The second part is dedicated to a numerical procedure based on relaxation for additive kernel parameter estimation.
Finally, the efficiency of the proposed method is illustrated and compared to other approaches on Sobol's g-function.
\end{abstract}

\paragraph{keyword}
Kriging, Computer Experiment, Additive Models, GAM, Maximum Likelihood Estimation, Relaxed Optimization, Sensitivity Analysis

\maketitle

\newpage

\section{Introduction}

The study of numerical simulators often deals with calculation intensive computer codes. This cost implies that the number of evaluations of the numerical simulator is limited and thus many methods such as uncertainty propagation, sensitivity analysis, or global optimization are unaffordable. A well known approach to circumvent time limitations is to replace the numerical simulator by a mathematical approximation called metamodel (or response surface or surrogate model) based on the responses of the simulator for a limited number of inputs called the Design of Experiments (DoE). There is a large number of metamodels types and among the most popular we can cite regression, splines, neural networks... In this article, we focus on a particular type of metamodel: the Kriging method, more recently referred to as Gaussian Process modeling \cite{Rasmussen2006}. 
Originally presented in spatial statistics \cite{Cressie1993} as an optimal Linear Unbiased Predictor (LUP) of random processes, Kriging has become very popular in machine learning, where its interpretation is usually restricted to the convenient framework of Gaussian Processes (GP). Beyond the LUP ---which then elegantly coincides with a conditional expectation---, the latter GP interpretation allows indeed the explicit derivation of conditional probability distributions for the response values at any point or set of points in the input space. 

\medskip

The classical Kriging method faces two issues when the number of dimensions $d$ of the input space $D\subset \mathbb{R}^{d}$ becomes high. Since this method is based on neighborhoods, it requires an increasing number of points in the DoE to cover the domain $D$. The second issue is that the number of anisotropic kernel parameters to be estimated increases with $d$ so that the estimation becomes particularly difficult for high dimensional input spaces \cite{K-T.Fang2006,OHagan2006}. 
An approach to get around the first issue is to consider specific features lowering complexity such as the family of Additive Models. In this case, the response can approximately be decomposed as a sum of univariate functions:
\begin{equation}
f(\mathbf{x}) = \mu + \sum_{i=1}^{d}{f_i(x_i)},
\label{eq:AM}
\end{equation}
where $\mu \in \mathbb{R}$ and the $f_i$'s may be non-linear. Since their introduction by Stones in 1985 \cite{stone1985additive}, many methods have been proposed for the estimation of additive models. We can cite the method of marginal integration \cite{newey1994kernel} and a very popular method described by Hastie and Tibshirani in \cite{Buja1989,Hastie1990}: the GAM backfitting algorithm. However, those methods do not consider the probabilistic framework of GP modeling and do not usually provide additional information such as the prediction variance. Combining the high-dimensional advantages of GAMs with the versatility of GPs is the main goal of the present work. For the study functions that contain an additive part plus a limited number of interactions, an extension of the present work can be found in the recent paper of T. Muehlenstaedt~\cite{Muhl}.\medskip

The first part of this paper focuses on the case of additive Gaussian Processes, their associated kernels and the properties of associated additive kriging models. The second part deals with a Relaxed Likelihood Maximization (RLM) procedure for the estimation of kernel parameters for Additive Kriging models. 
Finally, the proposed algorithm is compared with existing methods on a well known test function: the Sobol's g-function \cite{Saltelli2000}. It is shown within the latter example that Additive Kriging with RLM outperforms standard Kriging and produce similar performances as GAM. Due to its approximation performance and its built-in probabilistic framework both demonstrated later in this article, the proposed Additive Kriging model appears as a serious and promising challenger among additive models.

\section{Towards Additive Kriging}

\subsection{Additive random processes}

Lets first introduce the mathematical construction of an additive GP. A function $f:D \subset \mathbb{R}^d \longrightarrow \mathbb{R}$ is additive when it can be written $f(\mathbf{x}) = \sum_{i=1}^{d}{f_i(x_i)}$, where $x_i$ is the $i$-th component of the $d$-dimensional input vector $\mathbf{x}$ and the $f_i$'s are arbitrary univariate functions.
Let us first consider two independent real-valued first order stationary processes $Z_1$ and $Z_2$ defined over the same probability space $(\Omega, \mathcal{F},P)$ and indexed by $\mathbb{R}$, so that their trajectories $Z_i(.;\omega): x \in \mathbb{R} \longrightarrow Z_i(x;\omega)$ are univariate real-valued functions. 
Let $K_i:\mathbb{R}\times\mathbb{R} \longrightarrow \mathbb{R}$ be their respective covariance kernels and $\mu_1, \mu_2 \in \mathbb{R}$ their means. 
Then, the process $Z:=Z_{1}+Z_{2}$ defined over $(\Omega, \mathcal{F},P)$ and indexed by $\mathbb{R}^{2}$, and so that 
\begin{equation}
\forall \omega \in \Omega \ \forall \mathbf{x} \in \mathbb{R}^{2} \ Z(\mathbf{x};\omega)=Z_{1}(x_1;\omega)+Z_{2}(x_2;\omega),
\label{eq_sum_proc}
\end{equation} 
\noindent
has mean $\mu = \mu_1 + \mu_2$ and kernel $K(\mathbf{x},\mathbf{y})=K_1(x_1,y_1)+K_2(x_2,y_2)$. 
Following equation~\ref{eq_sum_proc}, the latter sum process clearly has additive paths. 
In this document, we call additive any kernel of the form $K:(\mathbf{x},\mathbf{y})\in \mathbb{R}^d \times \mathbb{R}^d \longrightarrow K(\mathbf{x},\mathbf{y})=\sum_{i=1}^{d}{K_i(x_i,y_i)}$ where the $K_i$'s are semi-positive definite (s.p.d.) symmetric kernels over $\mathbb{R}\times\mathbb{R}$. Although not commonly encountered in practice, it is well known that such a combination of s.p.d kernels is also a s.p.d. kernel in the direct sum space \cite{Rasmussen2006}. Moreover, one can show that the paths of any random process with additive kernel are additive in a certain sens:

\begin{prop}
\label{addproc}
Any (square integrable) random process $Z_{\mathbf{x}}$ possessing an additive kernel is additive up to a modification. 
In essence, it means that there exists a process $A_{\mathbf{x}}$ which paths are all additive, and such that $\forall \mathbf{x} \in X,\ \mathbb{P}(Z_{\mathbf{x}}=A_{\mathbf{x}})=1$.
\end{prop}
The proof of this property is given in appendix for $d=2$. For $d=n$ the proof follows the same pattern but the notations are more cumbersome. 
Note that the class of additive processes is not actually limited to processes with additive kernels. For example, let us consider $Z_1$ and $Z_2$ two correlated Gaussian processes on $(\Omega, \mathcal{F},P)$ such that the couple $(Z_1,Z_2)$ is Gaussian. Then $Z_{1}(x_1)+Z_{2}(x_2)$ is also a Gaussian process with additive paths but its kernel is not additive. However, in the next section, the term additive process will always refer to GP with additive kernels. 

\subsection{Invertibility of covariance matrices}
\begin{figure}[!ht]%
\begin{center}
\includegraphics[width=0.9\textwidth]{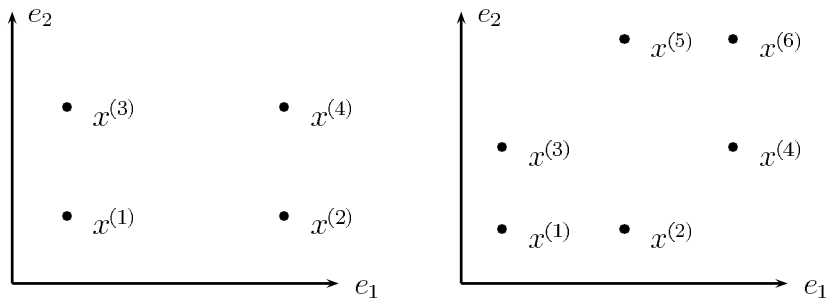}
\end{center}
\caption{2-dimensional examples of DoE which lead to non-invertible covariance matrix in the case of random processes with additive kernels.}%
\label{fig:planprob}
\end{figure}

In practice, the covariance matrix $\mathrm{K}$ of the observations of an additive process $Z$ at a design of experiments $X=(\mathbf{x}^{(1)}\ \dots\ \mathbf{x}^{(n)})^T$ may not be invertible even if there is no redundant point in $X$. Indeed, the additivity of $Z$ may introduce linear relationships (that holds almost surely) between the observed values of $Z$ and lead to the non invertibility of $\mathrm{K}$. Figure~\ref{fig:planprob} shows two examples of designs leading to a linear relationship between the observation. For the left panel, the additivity of $Z$ implies that $Z(\mathbf{x}^{(4)}) = Z(\mathbf{x}^{(2)})+Z(\mathbf{x}^{(3)})-Z(\mathbf{x}^{(1)})$ and thus the fourth column of the covariance matrix is a linear combination of the three other columns: $K(\mathbf{x}^{(i)},\mathbf{x}^{(4)})=K(\mathbf{x}^{(i)},\mathbf{x}^{(2)})+K(\mathbf{x}^{(i)},\mathbf{x}^{(3)})-K(\mathbf{x}^{(i)},\mathbf{x}^{(1)})$ and the associated covariance matrix is not invertible. 

\medskip

A first approach is to remove some points in order to avoid any linear combination, which is furthermore in accordance with the aim of parsimonious evaluations for costly simulators. Algebraic methods may be used for determining the combination of points leading to a linear relationship between the values of the random process but this procedure is out of the scope of this paper. 

\subsection{Additive Kriging}
Let $z : D \rightarrow \mathbb{R}$ be the function of interest (a numerical simulator for example), where $D \subset \mathbb{R}^d$. The responses of $z$ at the DoE $\mathcal{X}$ 
are noted $\mathbf{Z}=(z(\mathbf{x}^{(1)}) \ ... \ z(\mathbf{x}^{(n)}))^T$. Simple kriging relies on the hypothesis that $z$ is one path of a centered random process $Z$ with kernel $K$.  The expression of the best predictor (also called kriging mean) and of the prediction variance are :
\begin{equation}
\begin{split}
m(\mathbf{x}) & =  k(\mathbf{x})^T \mathrm{K} ^{-1}\mathbf{Z} \\
v(\mathbf{x}) & =  K(\mathbf{x},\mathbf{x}) - k(\mathbf{x})^T \mathrm{K} ^{-1}k(\mathbf{x})
\end{split}
\label{eq:BP}
\end{equation}
where $k(\mathbf{x})=(K(\mathbf{x},\mathbf{x}^{(1)})\ \dots\ K(\mathbf{x},\mathbf{x}^{(n)}))^T$ and $\mathrm{K}$ is the covariance matrix of general term $\mathrm{K}_{i,j}=K(\mathbf{x}^{(i)},\mathbf{x}^{(j)})$. 
Note that these equations respectively correspond to the conditional expectation and variance in the case of a GP with known kernel.
In practice, the structure of $k$ is supposed known (e.g. power-exponential or Mat\`ern families) but its parameters are unknown. A common way to estimate them is to maximize the likelihood of the kernel parameters given the observations $\mathbf{Z}$ \cite{Ginsbourger2009,Rasmussen2006}.
\medskip

\noindent
Equations \ref{eq:BP} are valid for any s.p.d kernel, thus they can be applied with additive kernels. In this case, the additivity of the kernel implies the additivity of the kriging mean: for example in dimension 2, for $K(\mathbf{x},\mathbf{y})=K_1(x_1,y_1)+K_2(x_2,y_2)$ we have
\begin{equation}
\begin{split}
m(\mathbf{x}) & =  (k_1(x_1)+k_2(x_2))^T(\mathrm{K_1+K_2})^{-1}\mathbf{Z}\\
& =  k_1(x_1)^T(\mathrm{K_1+K_2})^{-1}\mathbf{Z} + k_2(x_2)^T(\mathrm{K_1+K_2})^{-1}\mathbf{Z}\\
& = m_1(x_1) + m_2(x_2).
\end{split}
\end{equation}

\noindent
Another interesting property concerns the variance: $v$ can be null at points that do not belong to the DoE. Let us consider a two dimensional example where the DoE is composed of the 3 points represented on the left pannel of figure~\ref{fig:planprob}: $\mathcal{X}=\{\mathbf{x}^{(1)}\ \mathbf{x}^{(2)} \ \mathbf{x}^{(3)}\}$. Direct calculations presented in Appendix B shows that the prediction variance at the point $\mathbf{x}^{(4)}$ is equal to 0. 
This particularity follows from the fact that the value of the additive process are known almost surely at the point $x^{(4)}$ based on the observations at $\mathcal{X}$. In the next section, we illustrate the potential of Additive Kriging on an example and propose an algorithm for parameter estimation.

\subsection{Illustration and further consideration on a 2D example}
\begin{figure}[!ht]%
\begin{center}
\includegraphics[width=0.9\textwidth]{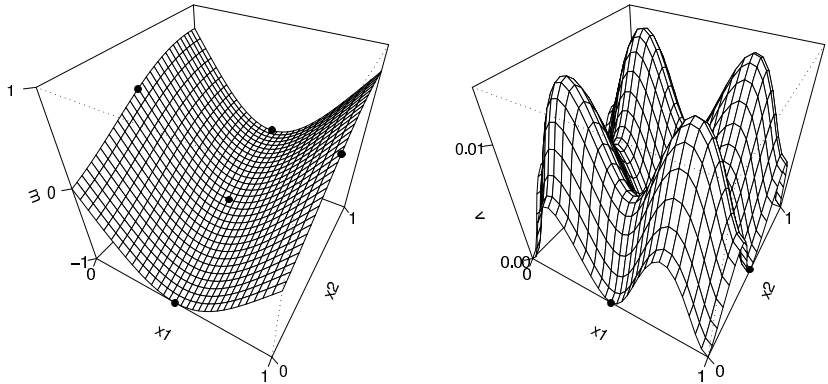}
\end{center}
\caption{Approximation of the function $f$ based on five observations (black dots). The left panel represents the best predictor and the right panel the prediction variance. the kernel used is the additive gaussian kernel with parameters $\sigma = (1\ 1)$ and $\theta = (0.6\ 0.6)$.}%
\label{fig:ex1dim2}
\end{figure}

We present here a first basic example of an additive kriging model. We consider $D=[0,1]^2$, and a set of 5 points in $D$ where the value of the observations $\mathbf{F}$ are arbitrarily chosen. Figure~\ref{fig:ex1dim2} shows the obtained kriging model. We can see on this figure the properties we mentioned above: the kriging mean is an additive function and the prediction variance can be null for points that do no belong to the DoE.

The effect of any variable can be isolated and represented so as the metamodel can be split in a sum of univariate sub-models. Moreover, we can get confidence intervals for each univariate model. As the expression of the first univariate model is
\begin{equation}
m_1(x_1)=k_1(x_1)^T(\mathrm{K_1}+\mathrm{K_2})^{-1}\mathbf{F}
\end{equation}
the effect of the direction 2 can be seen as an observation noise. We thus get an expression for the prediction variance of the first main effect
\begin{equation}
v_1(x_1)=K_1(x_1,x_1)-k_1(x_1)^T(\mathrm{K_1}+\mathrm{K_2})^{-1}k_1(x_1).
\end{equation}

\begin{figure}[!ht]%
\begin{center}
\includegraphics[width=0.9\textwidth]{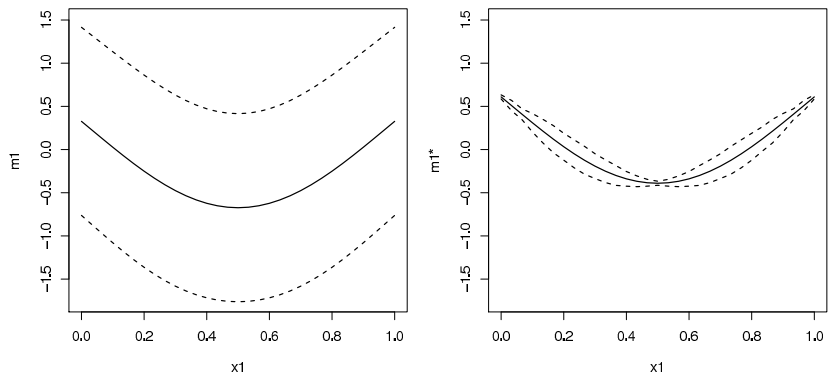}
\end{center}
\caption{Univariate models of the 2-dimensional example. The left panel plots $m_1$ and the 95\% confidence intervals $c_1(x_1)=m_1(x_1) \pm 2\sqrt{v_1(x_1)}$. The right panel shows the sub-model of the centrated univariate effects $m_1^*$ and $c_1^*(x_1)=m_1^*(x_1) \pm 2\sqrt{v_1^*(x_1)}$}%
\label{fig:ex1dim2bis}
\end{figure}
The left panel of figure~\ref{fig:ex1dim2bis} shows the obtained sub-model for the first direction. The interest of such graphic is limited since a 2-dimensional function can be plotted but this decomposition becomes useful to get an insight on the effect of a variable when $d>2$.  However, we can see that the confidence intervals are wide. This is because the sub-models are define up to a constant. In order to get rid of the effect of such a translation, an option is to approximate $Z_i(x_i)-\int{Z_i(s_i)\mathrm{d} s_i}$ conditionally to the observations:
\begin{equation}
\begin{split}
m_i^*(x_i)&=\E \left[ \left.Z_i(x_i)-\int{Z_i(s_i)\mathrm{d} s_i} \right| Z(X)=Y  \right] \\
v_i^*(x_i)&=\V \left[ \left.Z_i(x_i)-\int{Z_i(s_i)\mathrm{d} s_i} \right| Z(X)=Y  \right]
\end{split}
\end{equation}
The expression of $m_i^*(x_i)$ is straightforward whereas $v_i^*(x_i)$ requires more calculations which are given in Appendix C.
\begin{equation}
\begin{split}
m_i^*(x_i)&=m_i(x_i)-\int{m_i(s_i)\mathrm{d} s_i} \\
v_i^*(x_i)&= v_i(x_i) - 2 \int{K_i(x_i,s_i)\mathrm{d} s_i} + 2\int{k_i(x_i)^T K^{-1} k_i(s_i) \mathrm{d} s_i} \\
& \hspace{2cm} + \iint{K_i(s_i,t_i)\mathrm{d} s_i \mathrm{d} t_i} - \iint{k_i(t_i)^T K^{-1} k_i(s_i) \mathrm{d} s_i  \mathrm{d} t_i}
\end{split}
\end{equation}
The benefits of using $m_i^*$ and $v_i^*$ and then to define the sub-models up to a constant can be seen on the right panel of figure~\ref{fig:ex1dim2bis}. At the end, the probabilistic framework gives an insight on the error of the metamodel but also of each sub-model. 

\section{Parameter estimation} 

\subsection{Maximum likelihood estimation (MLE)}
MLE is a standard way to estimate covariance parameters and it is covered in detail in the literature \cite{Rasmussen2006,Santner2003}. Let $Y$ be a centered additive Process and $\psi_i=\{ \sigma_i^2, \theta_i \} \text{ with } i \in \{1,\dots,d\}$ the parameters of the univariate kernels. According to the MLE methodology, the best values $\psi_i^*$ for the parameters $\psi_i$ are the values maximizing the likelihood of the observations $\mathrm{Y}$:
\begin{equation}
\mathcal{L}(\psi_1,\dots,\psi_d) := \frac{1}{{(2\pi)}^{n/2} \mathrm{det(K(\psi))}^{1/2}} \mathrm{exp} \left( - \frac12 \mathrm{Y}^T \mathrm{K}(\psi)^{-1} \mathrm{Y} \right)
\label{eq:L}
\end{equation}
where $\mathrm{K}(\psi) = \mathrm{K}_1(\psi_1) + \dots +\mathrm{K}_d(\psi_d)$ is the covariance matrix depending on the parameters $\psi_i$. The latter maximization problem is equivalent to the usually preferred minimization of 
\begin{equation}
l(\psi_1,\dots,\psi_d) := \mathrm{log}(\mathrm{det}(\mathrm{K}(\psi))) + \mathrm{Y}^T \mathrm{K}(\psi)^{-1} \mathrm{Y}
\label{eq:RLL}
\end{equation} 
Obtaining the optimal parameters $\psi_i^*$ relies on the succesful use of a non-convex global optimization routine. 
This can be severely hindered for large values of $d$ since the search space of kernel parameters becomes high dimensional. One way to cope with this issue is to separate the variables and split the 
optimization into several low-dimensional subproblems.


\subsection{The Relaxed Likelihood Maximization algorithm}
The aim of the Relaxed Likelihood Maximization (RLM) algorithm is to treat separately the optimization in each direction. In this way, RLM can be seen as a cyclic relaxation optimization procedure \cite{Minoux1986} with initial values of the parameters $\sigma^2_i$ set to zero. 
As we will see, the main originality here is to consider a kriging model with an observation noise variance $\tau^2$ that fluctuates during the optimization. This parameter account for the metamodel error (if the function is not additive for example) but also for the inaccuracy of the intermediate values of $\sigma_i$ and $\theta_i$. \medskip

\noindent
The first step of the algorithm is to estimate the parameters of the kernel $K_1$. The simplification of the method is to consider that all the variations of $Y$ in the other directions can be summed up as a white noise. Under this hypothesis, $l$ depends on $\psi_1$ and 
$\tau$:
\begin{equation}
l(\psi_1,\tau) = \mathrm{log}(\mathrm{det}(\mathrm{K}_1(\psi_1)+\tau^2 I_d)) + \mathbf{Y}^T (\mathrm{K}_1(\psi_1)+\tau^2 I_d)^{-1} \mathbf{Y}
\label{eq:KAM-E1}
\end{equation}

\noindent
Then, the couple $\{\psi_1^*,\tau^{*} \}$ that maximizes $l(\psi_1,\tau)$ can be obtained by numerical optimization. 

\noindent
The second step of the algorithm consists in estimating $\psi_2$, with $\psi_1$ fixed to $\psi_1^*$: 
\begin{equation}
\begin{split}
\{ \psi_2^*, \tau^{*} \} = & \underset{\psi_2, \tau}{\mathrm{argmax}}(l(\psi_1^*,\psi_2,\tau)) \text{, with} \\
l(\psi_1^*,\psi_2,\tau) = & \mathrm{log}(\mathrm{det}(\mathrm{K}_1(\psi_1^*)+\mathrm{K}_2(\psi_2)+\tau^2 I_d)) + \\
 & \quad \mathbf{Y}^T (\mathrm{K}_1(\psi_1^*)+\mathrm{K}_2(\psi_2)+\tau^2 I_d)^{-1} \mathbf{Y}
\end{split}
\label{eq:KAM-E2}
\end{equation}

This operation can be repeated for all the directions until the estimation of $\psi_d$. 
\noindent
However, even if all the parameters $\psi_i$ have been estimated, it is fruitful to re-estimate them such that the estimation of the parameter $\psi_i$ can benefit of the values $\psi_j^*$ for $j>i$. Thus, the algorithm is composed of a cycle of estimations that treat each direction one after each other: 
\bigskip

\noindent
\textbf{RLM Algorithm :}\\
 1. Initialize the values $\sigma_i^{(0)}=0$ for $\ i \in \{1,\dots,d \}$ \\
 2. For $k$ from 1 to number of iteration do \\
 3. \quad For $l$ from 1 to $d$ do \\
 4. \qquad $ \{ \psi_l^{(k)},\tau^{(k)} \}= \underset{\psi_l,\tau}{\mathrm{argmin}}(l_c(\psi_1^{(k)},\dots,\psi_{l-1}^{(k)},\psi_{l},\psi_{l+1}^{(k-1)},\dots,\psi_{d}^{(k-1)},\tau))$ \\
 5. \quad   End For \\
 6. End For

\bigskip

\noindent
$\tau$ is a parameter tuning the fidelity of the metamodel since for $\tau=0$ the kriging mean interpolates the data. In practice, this parameter is decreasing at almost each new estimation. Depending on the observations and on the DoE, $\tau$ converges either to a constant or to zero (cf. the g-function example and figure~\ref{fig:dim4pep}). When zero is not reached, $\tau^2$ should correspond to the part of the variance that cannot be explained by the additive model. Thus, the comparison between $\tau^2$ and the $\sigma_i^2$ allows us to quantify the degree of additivity of the objective function according to the metamodel.
\medskip

\noindent
This procedure of estimation is not meant to be applied for kernels that are not additive. The method developed by Welch for tensor product kernels in \cite{WELCH1992} has similarities since it corresponds to a sequential estimation of the parameters. One interesting feature of Welch's algorithm is to choose at each step the best search direction for the parameters. The RLM algorithm could easily be adapted in a similar way to improve the quality of the results but the corresponding adapted version would be much more time consuming. 

\section{Comparison between the optimization's methods}
The aim of this section is to compare the RLM algorithm to the Usual Likelihood Maximization (ULM). The test functions that are considered are paths of an additive GP $Y$ with Gaussian additive kernel $K$. For this example, the parameters of $K$ are fixed to $\sigma_i=1$, $\theta_i=0.2$ for $i \in 1 \dots d$ but those values are supposed to be unknown.

\noindent
Here, $2 \times d+1$ parameters have to be estimated: $d$ for the variances, $d$ for the range and 1 for the noise variance $\tau^2$. For ULM, they are estimated simultaneously, whereas the RLM is a 3-dimensional optimization at each step. In both cases, we use the \verb?L-BFGS-B? method of the function \verb?optim? with the \verb?R? software. To judge the effectiveness of the algorithms, we compare here the best value found for the log-likelihood $l$ to the computational budget (the number of call to $l$) required for the optimization. As the \verb?optim? function returns the number $nc$ of call to $l$ and the best value $bv$ at the end of each optimization, we obtain for the MLE on one path of $Y$ one value of $nc$ and $bv$ for ULM and $nb\_iteration \times d$ values of $nv$ and $bv$ for RLM since there is one optimization at each step of each iteration.

\begin{figure}[ht]
\centering
\subfigure[$d=10$]{
\includegraphics[width=5.5cm]{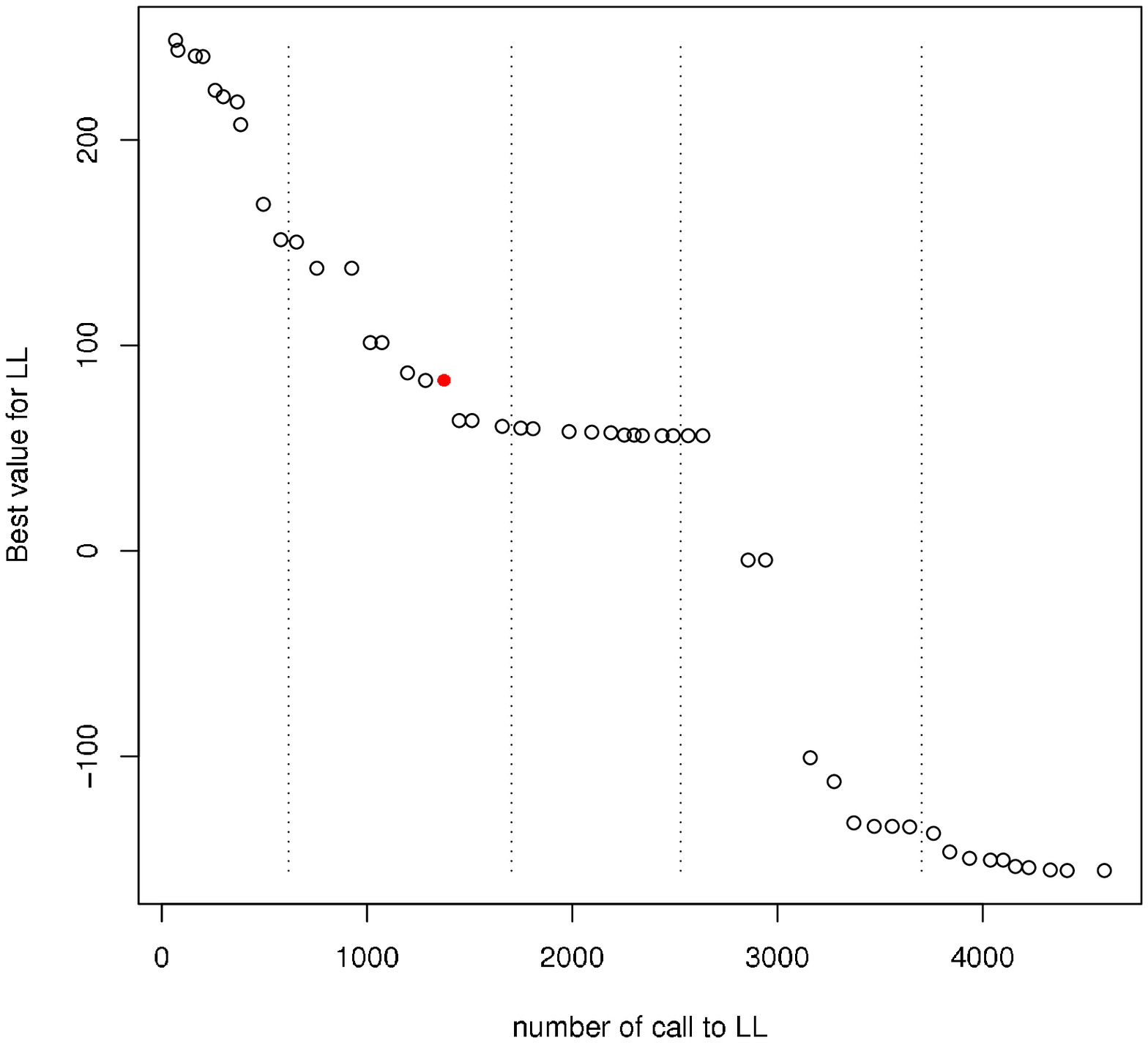}
}
\subfigure[$d=3$]{
\includegraphics[width=5.5cm]{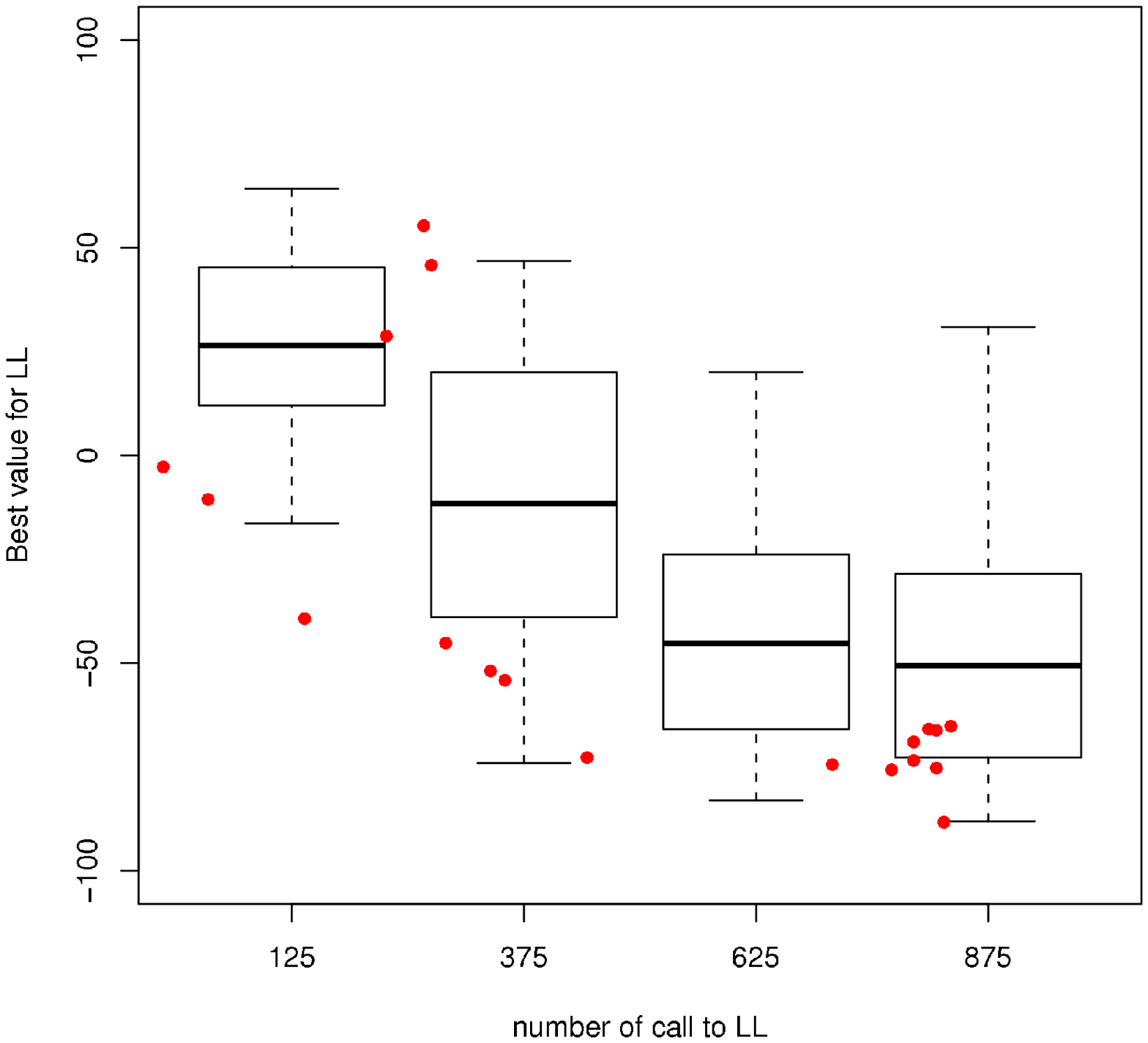}
}
\subfigure[$d=9$]{
\includegraphics[width=5.5cm]{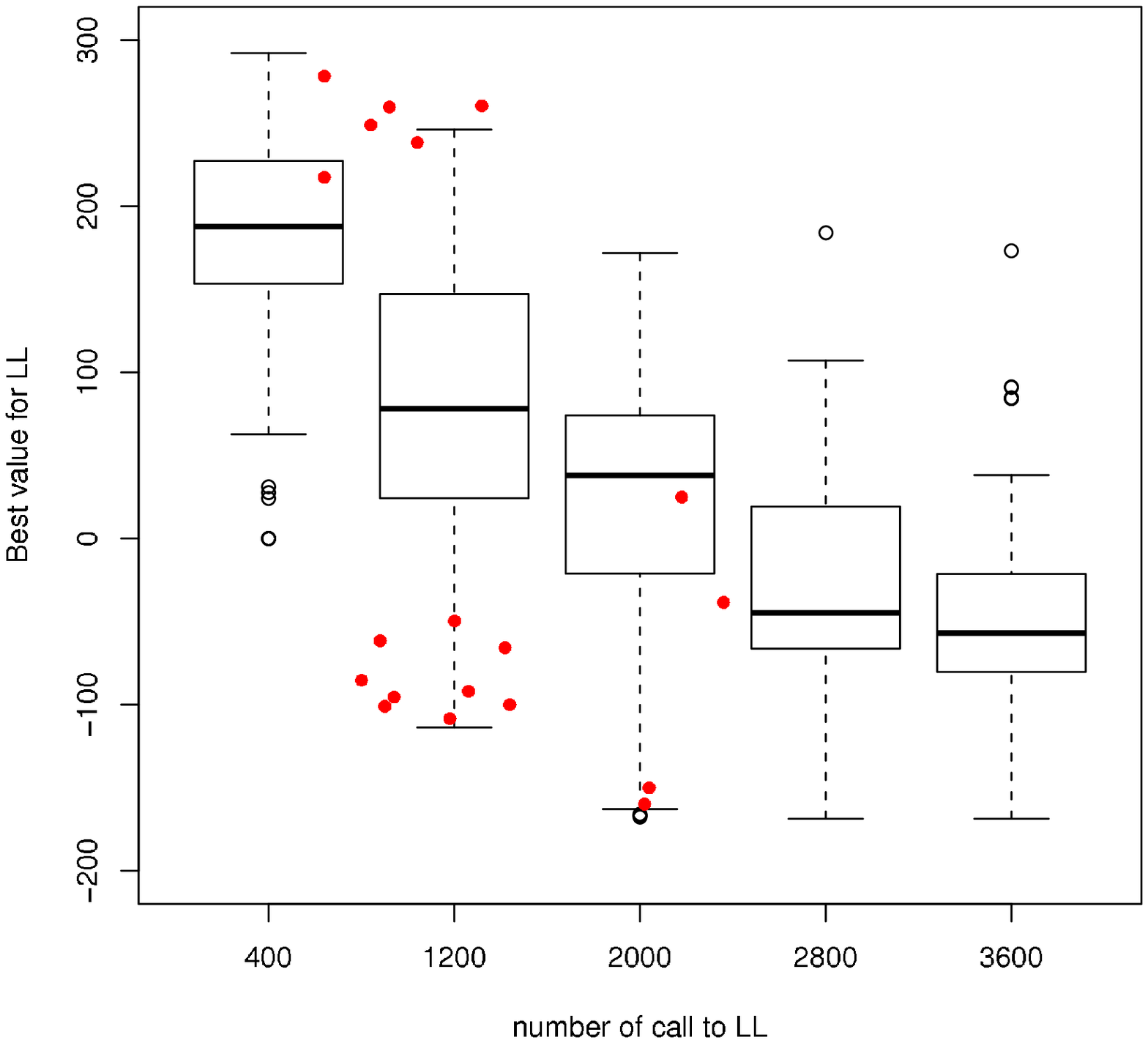}
}
\subfigure[$d=15$]{
\includegraphics[width=5.5cm]{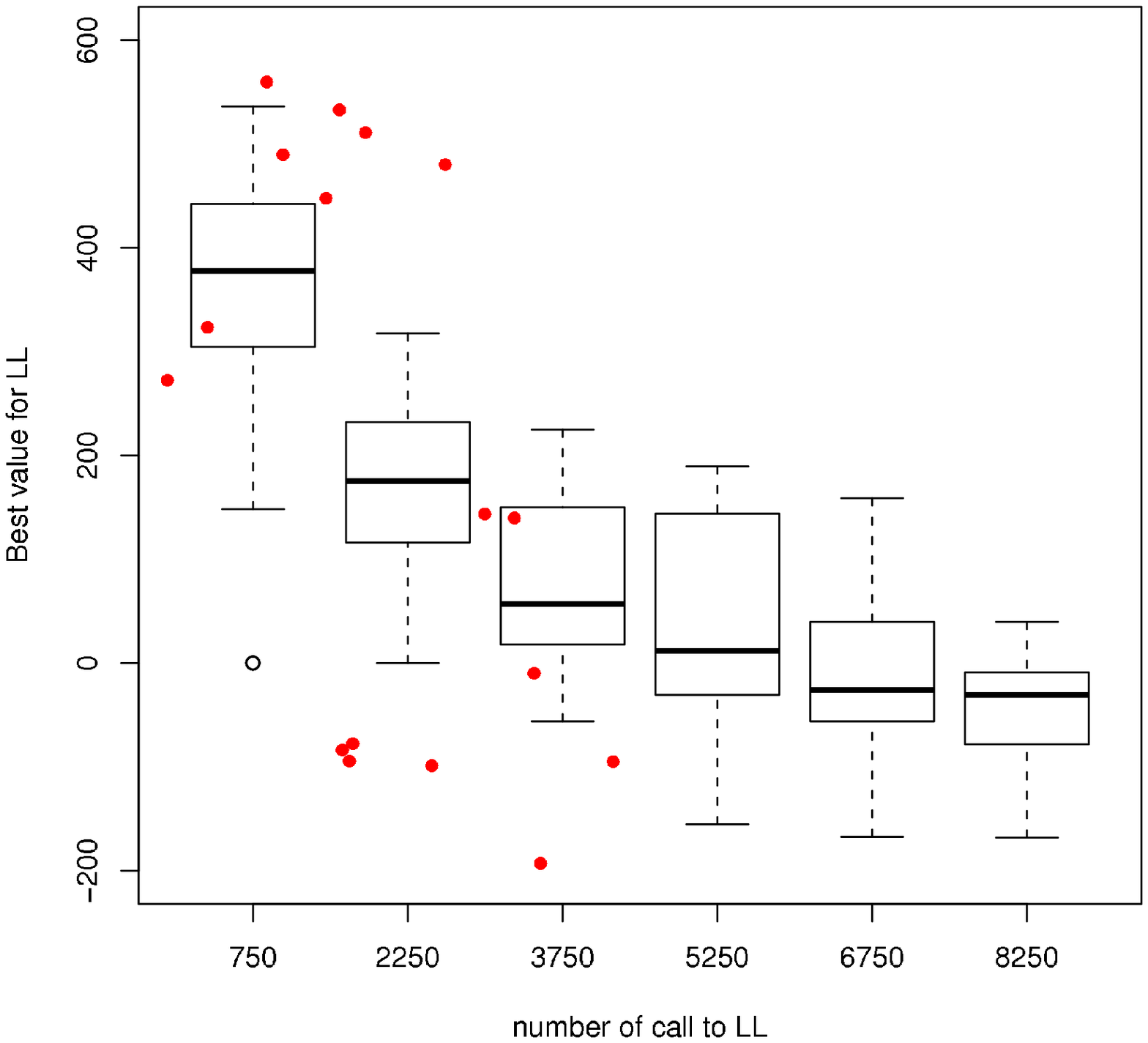}
}
\caption[Optional caption for list of figures]{Comparison of the optimization methods. The solid red dots are for usual optimizations and the black points are for the RLM method. For (a), the vertical lines correspond to the limit between two iterations of the algorithm. For (b), (c) and (d), the boxplots are based on 20 paths of $Y$ and each one sum up the best values of $l$ for a given range of number of call.}
\label{fig:COMP}
\end{figure}

\begin{figure}[ht]
\centering
\subfigure[$d=3$]{
\includegraphics[width=2.5cm]{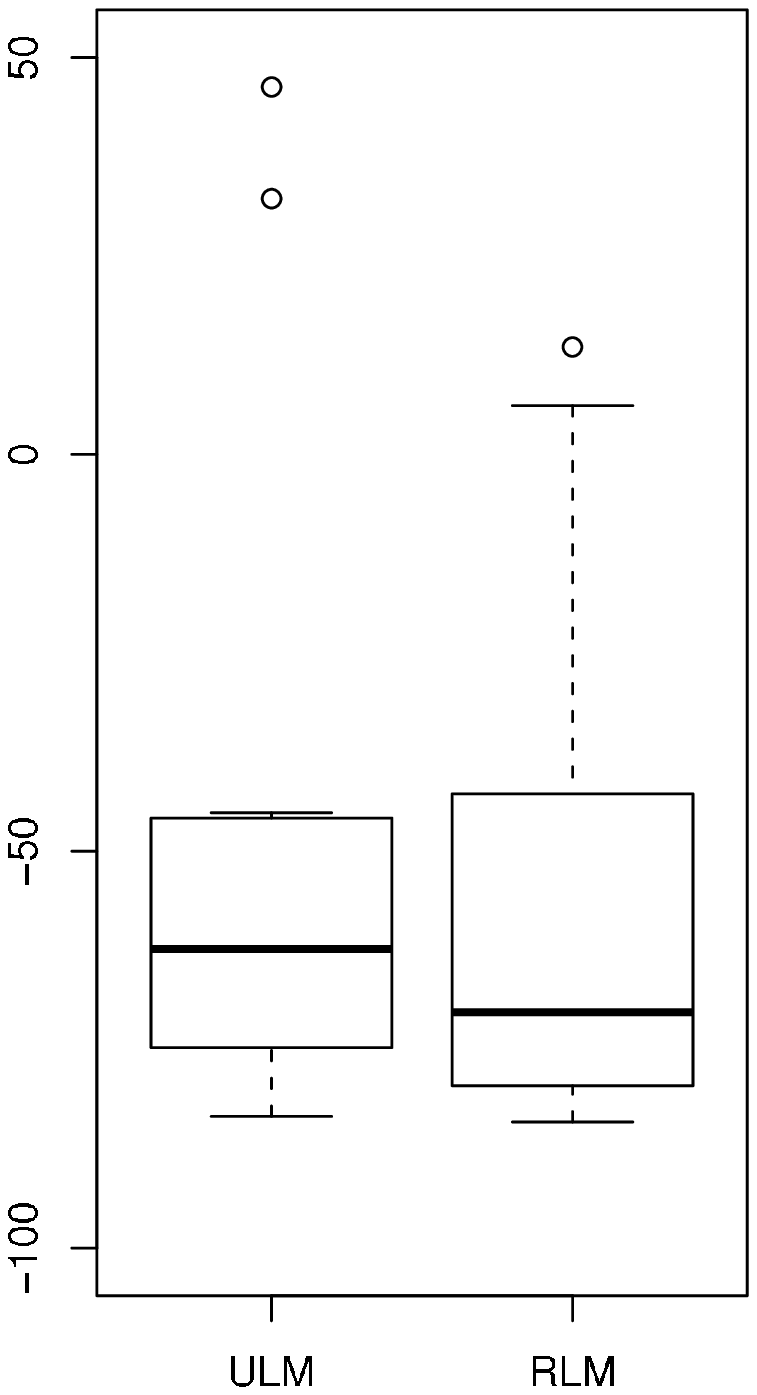}
}
\subfigure[$d=6$]{
\includegraphics[width=2.5cm]{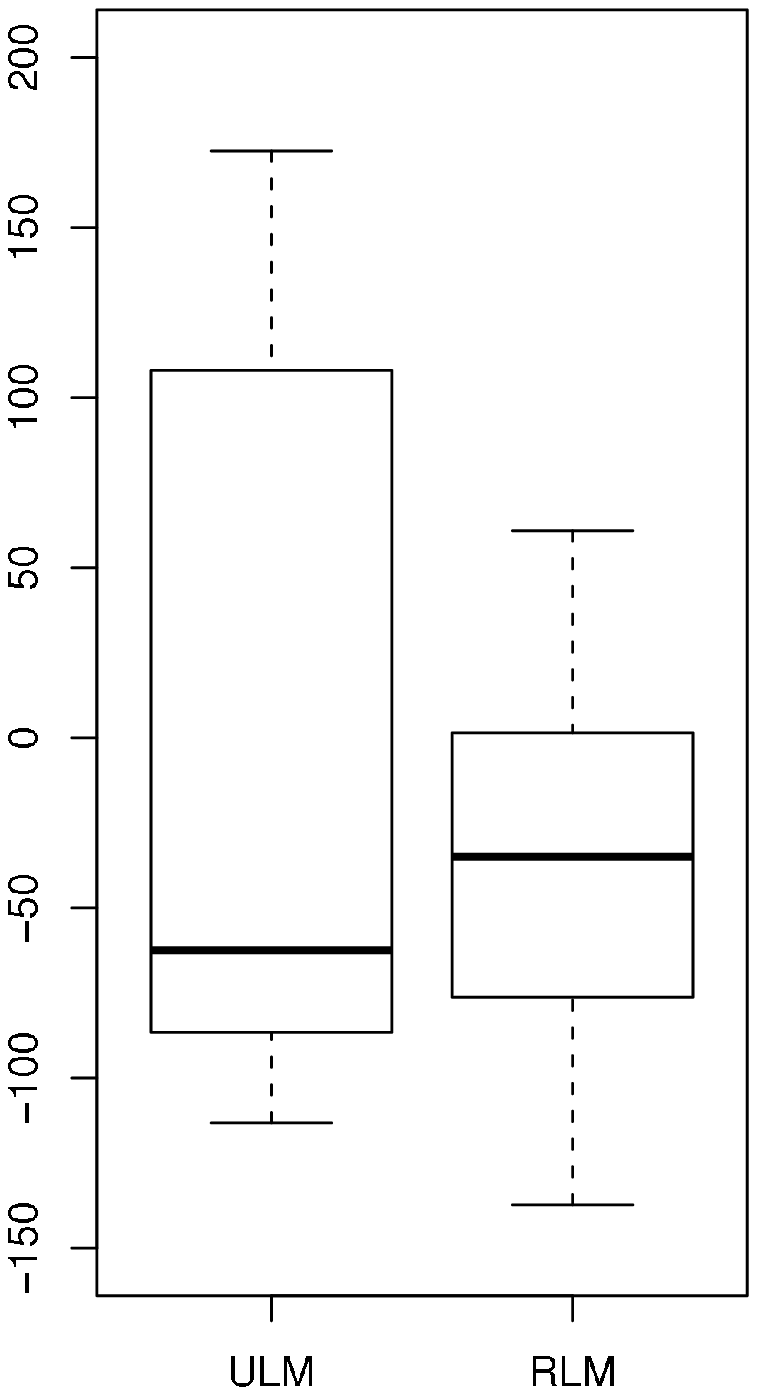}
}
\subfigure[$d=12$]{
\includegraphics[width=2.5cm]{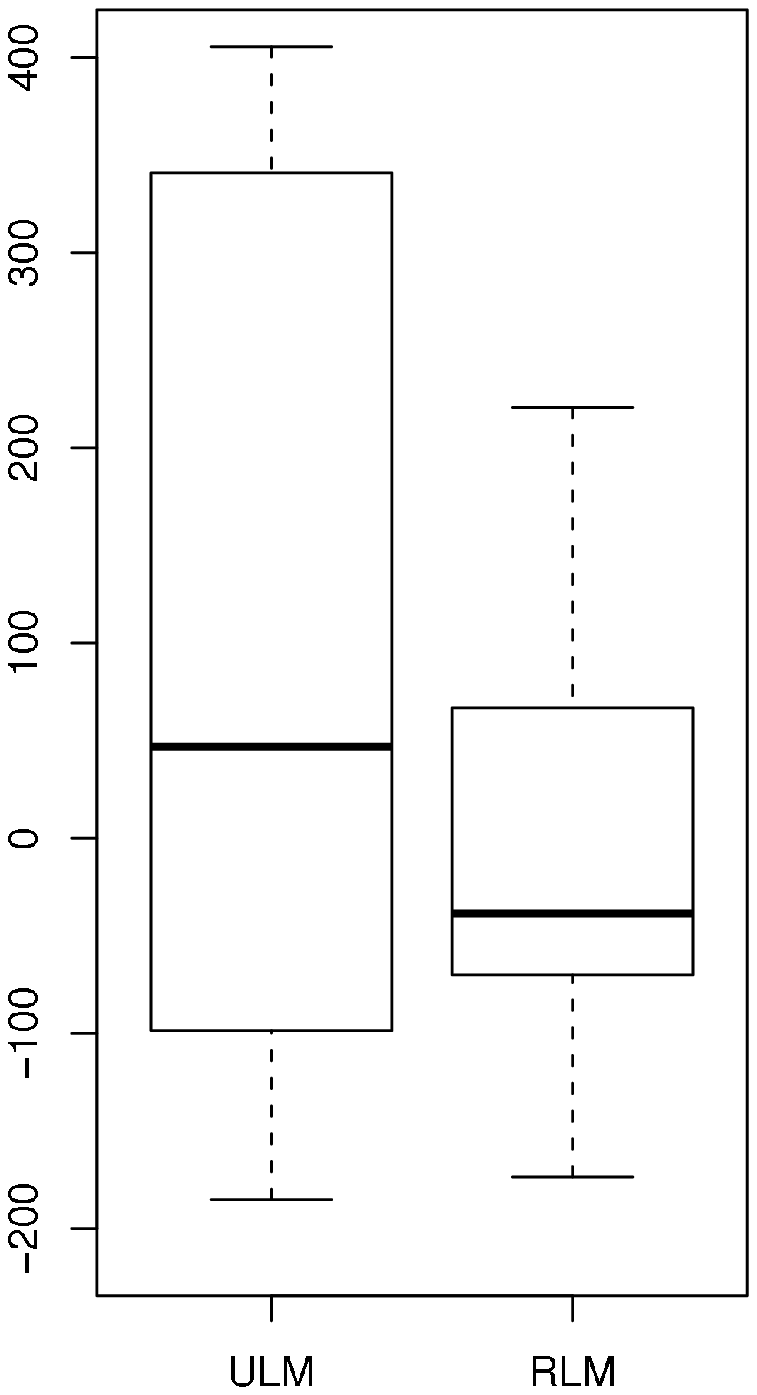}
}
\subfigure[$d=18$]{
\includegraphics[width=2.5cm]{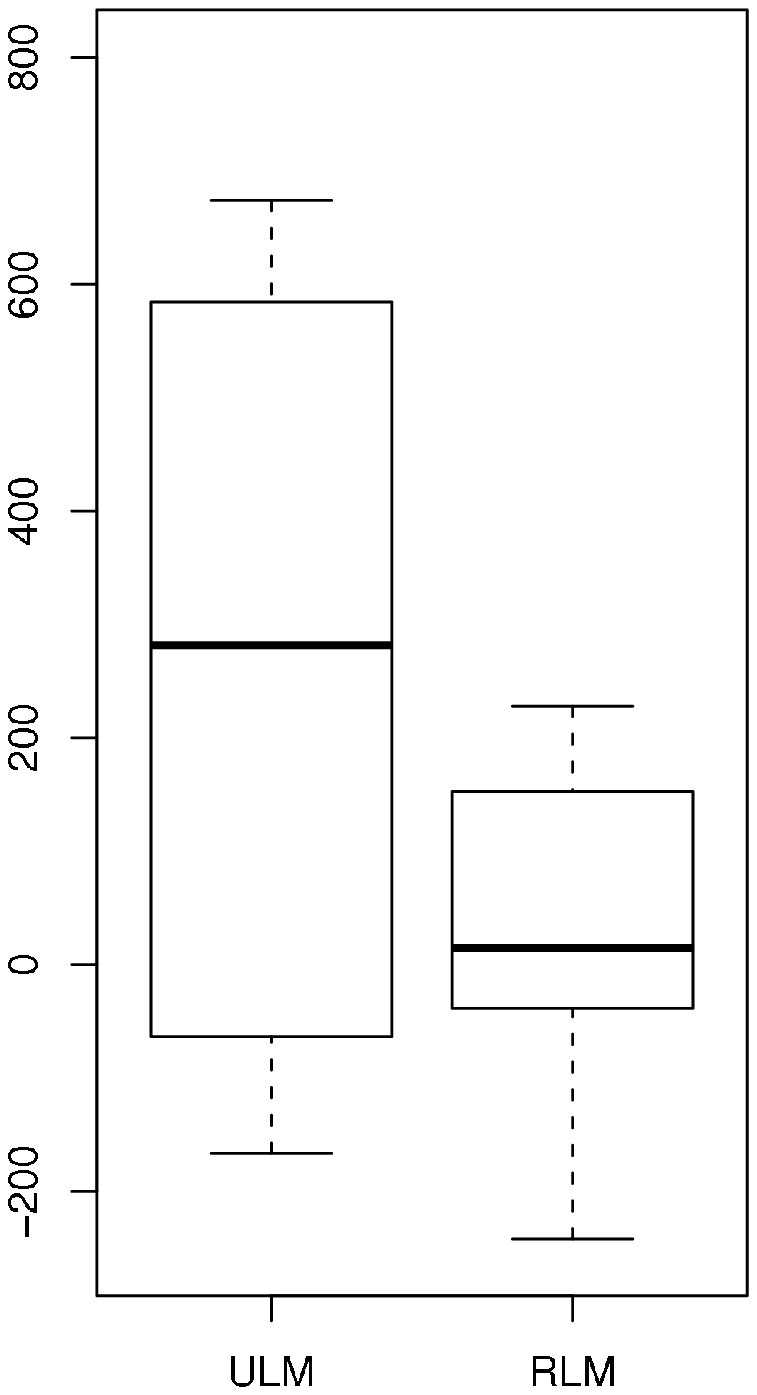}
}
\caption{Comparison of the final value for the log-likelihood with the two optimization algorithms for 20 paths of $Y$.}
\label{fig:COM2}
\end{figure}

\noindent
The panel (a) of figure~\ref{fig:COMP} presents the results for the two optimizations on a path of a GP for $d=5$. On this example, we can see that ULM needs 1500 calls to the log-likelihood before convergence whereas RLM requires much more calls before convergence. However, the result of the two methods are similar for 1500 calls but the result of RLM after 5000 calls is substantially improved. In order to get more robust results we simulate 20 paths of $Y$ and  we observe the global distribution of the variables $nv$ and $bv$. Furthermore, we study the evolution of the algorithm performances when the dimension increases choosing various values for the parameter $d$ from 3 to 18 with a Latin Hypercube (LH) Design with maximin criteria~\cite{Santner2003} containing $10\times d$ points. We observe on the panels (b), (c) and (d) of  figure~\ref{fig:COMP} that optimization with the RLM requires more calls to the function $l$, but this method appears to be more efficient and robust than ULM. Those results are stressed by figure~\ref{fig:COM2} where the final best value of RLM and ULM are compared. This figure also shows that the advantage of using RLM comes bigger when $d$ is getting larger.

\section{Application to the g-function of Sobol}
In order to illustrate the methodology and to compare it to existing algorithms, an analytical test case is considered. The function to approximate 
is the g-function of Sobol defined over $[0,1]^d$ by
\begin{equation}
g(\mathbf{x})= \prod_{k=1}^d \frac{|4x_k-2|+a_k}{1+a_k} \text{   with } a_k > 0
\label{eq:Gsob}
\end{equation}
This popular function in the literature \cite{Saltelli2000} is obviously not additive. However, depending on the coefficients $a_k$, $g$ can be very close to an additive function. As a rule, the g-function is all the more additive as the $a_k$ are large. One main advantage for our study is that the Sobol sensitivity indices can be obtained analytically so we can quantify the degree of additivity of the test function. For $i=1,\dots,d$ the indice $S_i$ associated to the variables $x_i$ is
\begin{equation}
S_i = \frac{\frac{1}{3(1+a_i)^2}}{\left[\prod_{k=1}^{d}1+\frac{1}{3(1+a_k)^2}\right]-1}.
\label{eq:SI}
\end{equation}

\noindent
Here we limit ourselves to the case $d=4$ and following \cite{Marrel2008} we choose $a_k = k$ for $k\in\{1,...,4\}$. For this combination of parameters, the sum of the first order Sobol indices is 0.95 so the g-function is almost additive. The considered DoE are LH maximin designs based on 40 points. To asses the quality of the obtained metamodels, the predictivity coefficient $Q_2$ is computed on a test sample of $n=1000$ points uniformly distributed over $[0,1]^4$. Its expression is:
\begin{equation}
Q_2(\mathbf{y},\hat{\mathbf{y}}) = 1 - \frac{\sum_{i=1}^{n}(y_i-\hat{y}_i)^2}{\sum_{i=1}^{n}(y_i-\bar{\mathbf{y}})^2}
\label{eq:Q2}
\end{equation}
where $\mathbf{y}$ is the vector of the values at the test points, $\hat{\mathbf{y}}$ is the vector of predicted values and $\bar{\mathbf{y}}$ is the mean of $\mathbf{y}$.

\begin{figure}[!ht]%
\begin{center}
\includegraphics[width=0.9\textwidth]{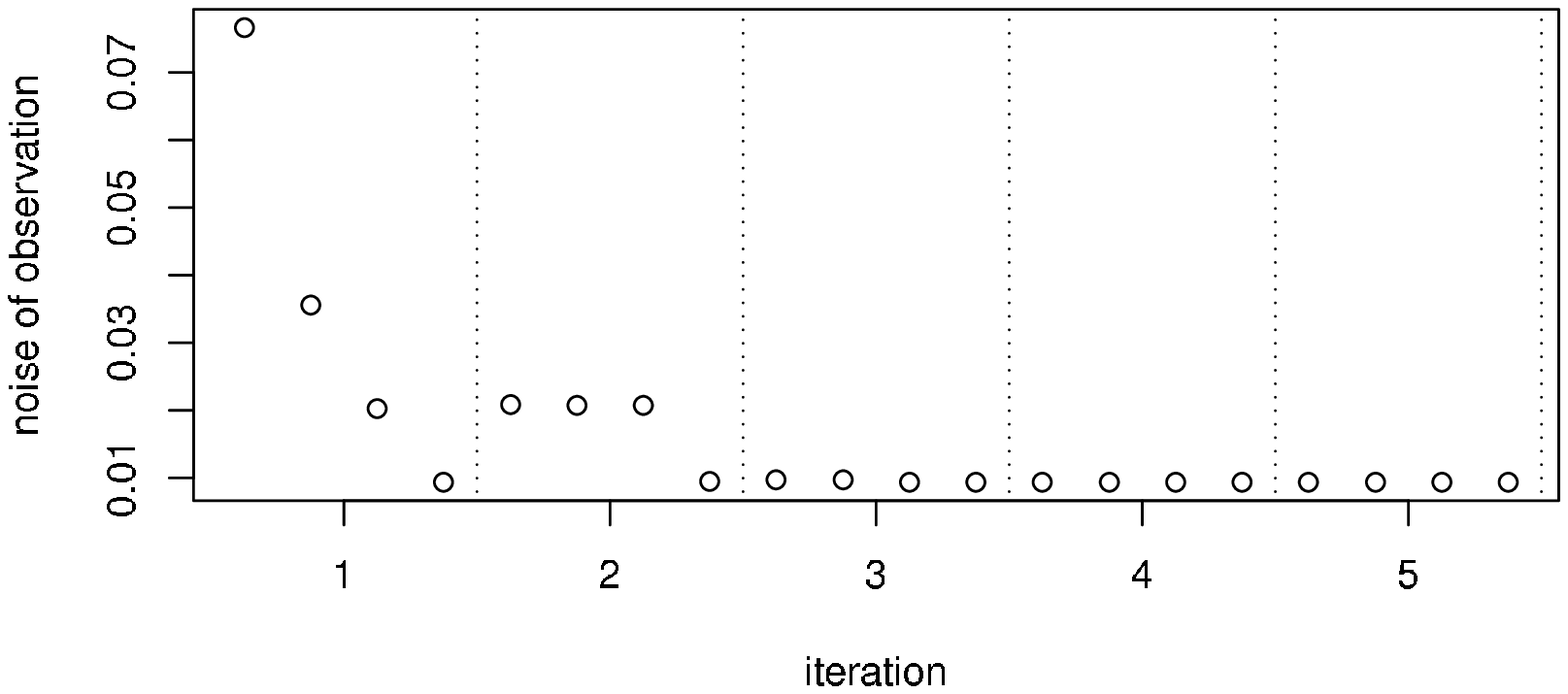}
\end{center}
\caption{Evolution of the observation noise on the 4-dimensional example}%
\label{fig:dim4pep}
\end{figure}

\noindent
We run on this example 5 iterations of the RLM algorithm with kernel Mat\`ern 3/2. The evolution of the estimated observation noise $\tau^2$ is represented on figure~\ref{fig:dim4pep}. On this figure, it appears that the observation noise is decreasing as the estimation of the parameters is improved. Here, the convergence of the algorithm is reached at iteration 4. The overall quality of the constructed metamodel is high since $Q_2=0.91$ and the final value for $\tau^2$ is 0.01.

\begin{figure}[!ht]%
\begin{center}
\includegraphics[width=0.9\textwidth]{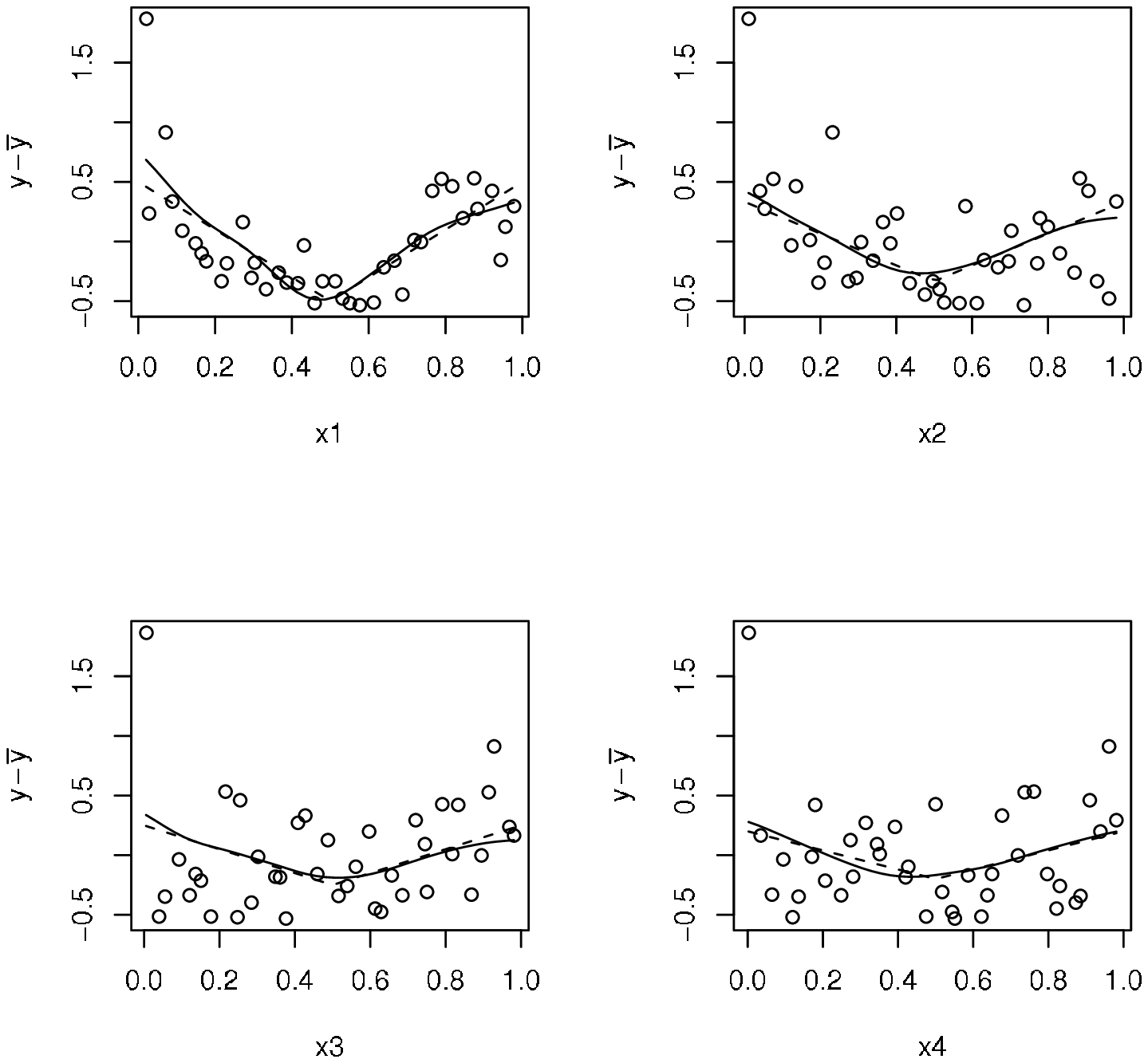}
\end{center}
\caption{$1$-dimensional projections of the observations (bullets) on the g-function example for $d=4$. 
The univariate models (solid lines) obtained after 5 iterations of RLM are very closed to the analytical main effects (dashed lines).}%
\label{fig:dim4dir}
\end{figure}

\medskip
\noindent
As previously the expression of the univariate sub-metamodels is
\begin{equation}
m_i(x_i)=k_i(x_i)^T(\mathrm{K_1}+\mathrm{K_2}+\mathrm{K_3}+\mathrm{K_4})^{-1}\mathbf{Y}
\end{equation}
The univariate functions obtained are presented on figure~\ref{fig:dim4dir}. The confidence intervals are not represented here in order to enhance the readability of the graphics and the represented values are centered to ensure that the observations and the univariate functions are comparable.

\medskip
\noindent
As the value of $Q_2$ is likely to fluctuate with the DoE and the optimization performances, we compare here the proposed RLM algorithm with other methods for 20 different LHS. The other methods used for the test are (a) additive kriging model with ULM, (b) kriging with usual tensor-product kernel, (c) the GAM algorithm. The results for classical kriging and GAM are obtained with the DiceKriging\footnote{As for RLM and ULM, DiceKriging also use the BFGS algorithm for the likelihood maximization} \cite{Roustant2009} and the GAM packages for \verb?R? available on the CRAN~\cite{RR}. As the value of the $a_k$ are the same as in \cite{Marrel2008} where Marrel et al. presents a specific algorithm for sequential parameter estimation in non-additive kriging models, the results of this paper are also presented as method (d). The mean and the standard deviation of the obtained $Q_2$ are gathered in table~\ref{tab:results}.
\begin{table}[h!]                                                                                                             
\centering                                                                                                                    
\begin{tabular}{|lc|cc|}                                                                                                     
\hline                                                                                                                        
Algorithm   & kernel               & $\mathrm{mean}(Q_2)$ & $\mathrm{sd}(Q_2)$     \\ \hline                                                             
RLM         & Additive Mat\`ern 3/2  & 0.90 & 0.016  \\                                                                
ULM         & Additive Mat\`ern 3/2  & 0.88 & 0.037  \\                                                     
Standart Kriging & Matern 3/2           & 0.82 & 0.042  \\                                                              
GAM         & (smoothing splines)  & 0.90 & 0.021  \\
Marrel      & power-exponential    & 0.86  & 0.07  \\ \hline                                                   
\end{tabular}                                                                                                                 
\caption{$Q_2$ predictivity coefficients at a 1000-points test sample for the various methods.}  
\label{tab:results}
\end{table}                                                                                                                   

\medskip

\section{Concluding remarks}
The proposed methodology seems to be a good challenger for additive modeling. On the example of the GP paths, the RLM appears to be more efficient than usual likelihood maximization and well suited for high dimensional modeling. On the second example, additive models benefit of the important additive component of the g-function and outperform non additive models even if the function is not purely additive. The predictivity of the RLM is equivalent to that of GAM but its robustness is higher for this example.

\medskip
\noindent
One main difference between RLM and GAM backfitting is that RLM takes into account the estimated parameters into the covariance structure whereas GAM subtracts from the observation the predicted value for all the sub-models obtained in the other directions. 

\medskip
\noindent
We would like to stress how the proposed metamodels take advantage of additivity, while benefiting from GP features. For the first point we can cite the complexity reduction and the interpretability. For the second, the main asset is that probabilistic metamodels provide the prediction variance. This justifies the fact of modeling an additive function on $\mathbb{R}^d$ instead of building $d$ metamodels over $\mathbb{R}$ since the prediction variance is not additive. We can also note the opportunity to choose a kernel adapted to the function to approximate. 

\medskip
\noindent
At the end, the proposed methodology is fully compatible with Kriging-based methods and its versatile applications. For example, one can choose a well suited kernel for the function to approximate or use additive kriging for high-dimensional optimization strategies relying on the expecting improvement criteria.

\bibliographystyle{elsarticle-num}

\appendix


\section*{Appendix A: Proof of proposition \ref{addproc} for $d=2$}
Let $Z$ be a random process indexed by $\mathbb{R}^2$ with kernel $K(\mathbf{x},\mathbf{y})= K_1(x_1,y_1) + K_2(x_2,y_2)$, and $Z_T$ the random process defined by $Z_T(x_1,x_2)=Z(x_1,0)+Z(0,x_2)-Z(0,0)$. By construction, the paths of $Z_T$ are additive functions. In order to show the additivity of the paths of $Z$, we will show that $\forall x \in \mathbb{R}^2$, $\p(Z(\mathbf{x})=Z_T(\mathbf{x}))=1$. For the sake of simplicity, the three terms of $\V[Z(\mathbf{x})-Z_T(\mathbf{x})]=\V[Z(\mathbf{x})]+\V[Z_T(\mathbf{x})]-2 \C[Z(\mathbf{x}),Z_T(\mathbf{x})]$ are studied separately:
\begin{equation*}
\V[Z(\mathbf{x})]=K(\mathbf{x},\mathbf{x})
\end{equation*}
\begin{equation*}
\begin{split}
\V[Z_T(\mathbf{x})] & = \V[Z(x_1,0)+Z(0,x_2)-Z(0,0)] \\
& = \V[Z(x_1,0)] + \V[Z(0,x_2)] + 2 \C[Z(x_1,0),Z(0,x_2)] \\
& \qquad + \V[Z(0,0)] - 2 \C[Z(x_1,0),Z(0,0)] - 2 \C[Z(0,x_2),Z(0,0)] \\
& = K_1(x_1,x_1) + K_2(0,0) + K_1(0,0) + K_2(x_2,x_2) + K(0,0) \\
& \qquad + 2 \left( K_1(x_1,0) + K_2(0,x_2)\right) - 2 \left( K_1(x_1,0) + K_2(0,0) \right) \\
& \qquad - 2 \left( K_1(0,0) + K_2(x_2,0) \right) \\
& = K_1(x_1,x_1) + K_2(x_2,x_2) = K(\mathbf{x},\mathbf{x})
\end{split}
\end{equation*}
\begin{equation*}
\begin{split}
\C[Z(\mathbf{x}),Z_T(\mathbf{x})] & = \C[Z(x_1,x_2),Z(x_1,0)+Z(0,x_2)-Z(0,0)] \\
& = K_1(x_1,x_1) + K_2(x_2,0) + K_1(x_1,0) + K_2(x_2,x_2) \\
& \qquad  - K_1(x_1,0) - K_2(x_2,0) \\
& = K_1(x_1,x_1) + K_2(x_2,x_2) = K(\mathbf{x},\mathbf{x})
\end{split}
\end{equation*}
Those three equations implies that $\V[Z(\mathbf{x})-Z_T(\mathbf{x})]=0$, $\forall{\mathbf{x}} \in \mathbb{R}^2$. Thus, $\p(Z(\mathbf{x})=Z_T(\mathbf{x}))=1$ and there exists a modification of $Z$ with additive paths.

\section*{Appendix B: Calculation of the prediction variance}
Let consider a DoE composed of the 3 points $\{\mathbf{x}^{(1)}\ \mathbf{x}^{(2)}\ \mathbf{x}^{(3)}\}$ represented on the left pannel of figure~\ref{fig:planprob}. We want here to show that although $\mathbf{x}^{(4)})$ does not belongs to the DoE we have $v(\mathbf{x}^{(4)})=0$.
\begin{eqnarray*}
v(\mathbf{x}^{(4)}) & = & K(\mathbf{x}^{(4)},\mathbf{x}^{(4)}) - k(\mathbf{x}^{(4)})^T \mathrm{K}^{-1} k(\mathbf{x}^{(4)}) \\
& = & K(\mathbf{x}^{(4)},\mathbf{x}^{(4)}) - (k(\mathbf{x}^{(2)})+k(\mathbf{x}^{(3)})-k(\mathbf{x}^{(1)}))^T \mathrm{K}^{-1} k(\mathbf{x}^{(4)}) \\
& = & K_1(\mathbf{x}^{(4)}_1,\mathbf{x}^{(4)}_1) + K_2(\mathbf{x}^{(4)}_2,\mathbf{x}^{(4)}_2) - \\
& & \quad(-1\ \ 1\ \ 1) 
\begin{pmatrix}
K_1(\mathbf{x}^{(1)}_1,\mathbf{x}^{(4)}_1)+K_2(\mathbf{x}^{(1)}_2,\mathbf{x}^{(4)}_2) \\
K_1(\mathbf{x}^{(2)}_1,\mathbf{x}^{(4)}_1)+K_2(\mathbf{x}^{(2)}_2,\mathbf{x}^{(4)}_2) \\
K_1(\mathbf{x}^{(3)}_1,\mathbf{x}^{(4)}_1)+K_2(\mathbf{x}^{(3)}_2,\mathbf{x}^{(4)}_2)
\end{pmatrix}
 \\
& = & K_1(\mathbf{x}^{(2)}_1,\mathbf{x}^{(2)}_1) + K_2(\mathbf{x}^{(3)}_2,\mathbf{x}^{(3)}_2) - K_1(\mathbf{x}^{(2)}_1,\mathbf{x}^{(2)}_1) - K_2(\mathbf{x}^{(3)}_2,\mathbf{x}^{(3)}_2) \\
& = 0
\end{eqnarray*}

\section*{Appendix C: Calculation of $v_i^*$}
We want here to calculate the variance of $Z_i(x_i)-\int{Z_i(s_i)\mathrm{d} s_i}$ conditionally to the observations $Y$.
\begin{equation*}
\begin{split}
v_i^*(x_i)&=\V \left[ \left.Z_i(x_i)-\int{Z_i(s_i)\mathrm{d} s_i} \right| Z(X)=Y  \right] \\
&=\V \left[ \left.Z_i(x_i) \right| Z(X)=Y  \right] - 2\C \left[ \left.Z_i(x_i),\int{Z_i(s_i)\mathrm{d} s_i} \right| Z(X)=Y  \right] \\
& \hspace{5cm} + \V \left[ \left.\int{Z_i(s_i)\mathrm{d} s_i} \right| Z(X)=Y  \right] \\
&= v_i(x_i) - 2 \left( \int{K_i(x_i,s_i)\mathrm{d} s_i} - \int{k_i(x_i)^T K^{-1} k_i(s_i) \mathrm{d} s_i} \right) \\
& \hspace{2cm} + \iint{K_i(s_i,t_i)\mathrm{d} s_i \mathrm{d} t_i} - \iint{k_i(t_i)^T K^{-1} k_i(s_i) \mathrm{d} s_i  \mathrm{d} t_i}
\end{split}
\end{equation*}

\end{document}